\def\year{2017}\relax
\documentclass[letterpaper]{article}
\usepackage{aaai17}
\usepackage{times}
\usepackage{helvet}
\usepackage{courier}

\usepackage{graphics} 
\usepackage{epsfig} 
\usepackage{mathptmx} 
\usepackage{amsmath} 
\usepackage{amssymb}  
\usepackage{algpseudocode}
\usepackage[textwidth=2cm,colorinlistoftodos]{todonotes}

\usepackage{textcase,microtype,marvosym}
\usepackage{amsmath,amssymb,graphicx,MnSymbol}
\usepackage{mathtools}
\usepackage{subfigure}
\usepackage{graphics} 
\usepackage{epsfig} 
\usepackage{nicefrac}
\usepackage{booktabs,colonequals}
\usepackage{comment}
\usepackage{tikz,pgfplots,pgfplotstable}
\pgfplotsset{compat=newest}
\pgfplotsset{plot coordinates/math parser=false}

\definecolor{lred}{RGB}{200,0,0} 
\definecolor{dred}{RGB}{130,0,0} 

\definecolor{dblu}{RGB}{0,0,130} 
\definecolor{dgre}{RGB}{0,130,0} 

\definecolor{dgra}{RGB}{50,50,50}
\definecolor{mgra}{RGB}{100,100,100}
\definecolor{lgra}{RGB}{220,220,220}
\definecolor{MPG}{RGB}{000,125,122}
\definecolor{ora}{RGB}{255,153,51}

\newcommand{\g}{\;|\;}

\newcommand{\N}{\mathcal{N}}

\renewcommand{\vec}{\boldsymbol}
\newcommand{\mat}{\boldsymbol}

\newcommand{\vw}{\vec{w}}

\newcommand{\vs}{\vec{s}}

\newcommand{\mA}{\mathbf{A}}
\newcommand{\mB}{\mathbf{B}}
\newcommand{\mC}{\mathbf{C}}
\newcommand{\mQ}{\mathbf{Q}}
\newcommand{\mR}{\mathbf{R}}

\newcommand{\mS}{\mathbf{S}}

\usetikzlibrary{arrows,shapes,plotmarks}

\tikzset{>=stealth'}
\tikzstyle{graphnode} =
   [circle,draw=black,minimum size=22pt,text centered,text
     width=22pt,inner sep=0pt]
\tikzstyle{var}   =[graphnode,fill=white]
\tikzstyle{obs}   =[graphnode,fill=black,text=white]
\tikzstyle{fac}   =[rectangle,draw=black,fill=black!25,minimum size=5pt]
\tikzstyle{facprior} =[rectangle,draw=black,fill=black,text=white,minimum size=5pt]
\tikzstyle{edge}  =[draw=white,double=black,thick,-]
\tikzstyle{prior} =[rectangle, draw=black, fill=black, minimum size=
5pt, inner sep=0pt]
\tikzstyle{dirprior} = [circle, draw=black, fill=black, minimum
size=5pt, inner sep=0pt]

\DeclareSymbolFont{stmry}{U}{stmry}{m}{n}
\DeclareMathSymbol\leftarrowtriangle\mathrel{stmry}{"5E}
\DeclareMathSymbol\rightarrowtriangle\mathrel{stmry}{"5F}

\begin{document}
%
\title{A Probabilistic Representation for Dynamic Movement Primitives}
\author{Franziska Meier$^{1,2}$ and Stefan Schaal$^{1,2}$\\
$^{1}$CLMC Lab, University of Southern California, Los Angeles, USA\\
$^{2}$ Autonomous Motion Department, MPI for Intelligent Systems, T\"ubingen, Germany
\thanks{This research was supported in part by National Science Foundation grants IIS-1205249, IIS-1017134, EECS-0926052, the Office of Naval Research, the Okawa Foundation, and the Max-Planck-Society.}
}
\maketitle

\begin{abstract}
  Dynamic Movement Primitives have successfully been used to realize
  imitation learning, trial-and-error learning, reinforcement
  learning, movement recognition and segmentation and control. Because
  of this they have become a popular representation for motor
  primitives. In this work, we showcase how DMPs can be reformulated
  as a probabilistic linear dynamical system with control inputs.
  Through this probabilistic representation of DMPs, algorithms such
  as Kalman filtering and smoothing are directly applicable to perform
  inference on proprioceptive sensor measurements during execution. We
  show that inference in this probabilistic model automatically leads
  to a feedback term to online modulate the execution of a DMP.
  Furthermore, we show how inference allows us to measure the
  likelihood that we are successfully executing a given motion
  primitive. In this context, we show initial results of using the
  probabilistic model to detect execution failures on a simulated
  movement primitive dataset.
\end{abstract}

\section{Introduction}
\label{sec:introduction}
One of the main challenges towards autonomous robots remains autonomous
motion generation. A key observation has been that in certain
environments, such as households, tasks that need to be executed tend
to contain very repetitive behaviors \cite{tenorth2009tum}. Thus, the
idea of identifying motion primitives to form building blocks for
motion generation has become very popular
\cite{Ijspeert2013,paraschos2013probabilistic}. A popular and
effective way to learn these motion primitives is through imitation
learning \cite{SchaalTCS1999}.
Part of the research in this area is concerned with motion
representation, and a variety of options have been proposed
\cite{Ijspeert2013,khansari2010imitation,Dragan:2015vq,Wilson:1999dk,paraschos2013probabilistic}.
However, recently, there has been interest in going beyond pure motion
representation. In the context of closing action-perception loops,
motor skill representations that allow to close that loop are
attractive. The ability to feed back sensory information during the
execution of a motion primitives promises a less error prone motion
execution framework \cite{Pastor2011}. Furthermore, the ability to
associate a memory of how a motor skill should feel during execution
can greatly enhance robustness in execution. This concept of enhancing
a motor skill representation with a sensory memory has been termed
associative skill memories (ASMs)\cite{Pastor_RAS_2012}.

The idea of associative skill memories lies in leveraging the
repeatability of motor skills to associate experienced sensor
information with the motion primitive. This allows to adapt previously
learned motor skills when this primitive is being executed again and
deviates from previously experienced sensor information
\cite{Pastor2011}. Thus far, existing approaches
\cite{Pastor2011,Gams_TransRob_2014} use the concept of coupling terms
to incorporate sensor feedback into DMPs which typically involves
setting task-dependent gains to modulate the effect of the feedback
term.

In this work we endow Dynamic Movement Primitives with a probabilistic
representation, that maintains their original functionality, but allows
for uncertainty propagation. Inference in this probabilistic DMP model
is realized through Kalman filtering and smoothing. An interesting
component of probabilistic DMPs is the fact that -- given a reference
signal to track -- sensor feedback is automatically considered when
executing a desired behavior. As a result, the feedback term is an
intrinsic part of this probabilistic formulation. Similar to Kalman
filtering, the feedback term is now scaled based on a uncertainty
based gain matrix. Besides this insight, we also highlight the
benefits of combining a probabilistic motion primitive representation
with a strong structural prior -- which is given by the DMP framework
in this work. We demonstrate the advantage of this by showing how to
make use of the probabilistic model to perform failure detection when
executing a motion primitive.

This paper is organized as follows: We start by reviewing the DMP
framework and related probabilistic motion representations in
the background section. Then in the main section we
propose a graphical model representation for DMPs, detail the learning
procedure and illustrate their usage. Finally, we discuss how of failure detection can be realized with probabilistic DMPs and evaluate it in that context.
\section{Background}
\label{sec:background}
Dynamic Movement Primitives (DMPs) encode a desired movement
trajectory in terms of the attractor dynamics of nonlinear
differential equations~\cite{Ijspeert2013}. For a 1 DOF system, the
equations are given as:
\begin{eqnarray}\label{eq:DMP}
\frac{1}{\tau} \dot{z} &=& \alpha_z (\beta_z(g - p) - z) + s f(x) \nonumber \\
\frac{1}{\tau} \dot{p} &=& z
\end{eqnarray}
such that $p,\dot{p},\ddot{p} = \dot{z}$ are position, velocity, and
acceleration of the movement trajectory, where
$$ f(x) = \frac{\sum \limits_{i=1}^N \psi_i w_i x}{\sum \limits_{i=1}^N \psi_i},  \; \text{with} \; \psi_i = \exp{\big( -h_i(x-c_i)^2 \big)}$$
with
$$ \frac{1}{\tau} \dot{x} = -\alpha_x x \; \text{ and } \; s= \frac{g-p_0}{g_{fit} - p_{0,fit}} = \frac{g-p_0}{\Delta g}$$
In general, it is assumed that the duration $\tau$ and goal position
$g$ are known. Thus, given $\tau$ and $g$ the DMP is parametrized
through weights $\mathbf{w}=(w_1,...,w_N)^T$ which are learned to
represent the shape of any smooth movement. During this fitting
process, the scaling variable $s$ is set to one, and the value of
$\Delta g$ is stored as a constant for the DMP.
\subsection{Probabilistic Motion Primitive Representations}
The benefits of taking a probabilistic approach to motion
representation has been discussed by a variety of authors
\cite{toussaint2009probabilistic,ruckert2013learned,Meier2011,Calinon:2012ca,paraschos2013probabilistic,khansari2011learning}.
The use of probabilistic models varies however. For instance, the
approaches presented in \cite{Calinon:2012ca,khansari2011learning}
take a dynamical systems view that utilizes statistical methods to
encode variability of the motion. In contrast,
\cite{toussaint2009probabilistic,ruckert2013learned} take a trajectory
optimization approach using a probabilistic planning system.
Here, as outlined in the introduction, we take the dynamical systems
view, as a first step towards an implementation of associate skill
memories. In previous work \cite{Meier2011} we have shown how we can
reformulate the DMP equations into a linear dynamical system
\cite{Bishop2006}. Inference and learning in this linear dynamical
system was formulated as a Kalman filtering/smoothing approach. This
Kalman filter view of DMPs allowed us to perform online movement
recognition \cite{Meier2011} and segmentation of complex motor skills
into underlying primitives \cite{meierAISTATS2012}.

Finally, compared to previous work on probabilistic motion primitives,
such as \cite{Calinon:2012ca,paraschos2013probabilistic}, this
representation explicitly represents the dynamical system structure
as a dynamic graphical model and adds the possibility of considering
sensor feedback as part of the inference process.
\section{Probabilistic Dynamic Movement Primitives}
\label{sec:prob_dmps}
In this section we introduce a new {\em Probabilistic Dynamic Movement
  Primitive} model. The goal of this work is to essentially provide a
probabilistic model that can replace the standard non-probabilistic
representation without loss in functionality. Thus, here we aim at
deriving a graphical model that explicitly maintains positions,
velocities and accelerations, such that a rollout of that model
creates a full desired trajectory.
\subsection{Deriving the Probabilistic DMP Model}
We start out by deriving the new formulation and then show how
learning of the new probabilistic DMPs is performed.
The transformation system of a 1-DOF DMP can be discretized via Euler
discretization, resulting in
\begin{align}
  \dot{z}_t &= \tau(\alpha_z(\beta_z(g-p_{t-1}) - z_{t-1}) +f )\label{eq:acc_level_1}\\
  z_t &= \dot{z}_t \Delta t + z_{t-1}\label{eq:vel_level_1}\\
  \ddot{p}_t &= \tau \dot{z}_t\label{eq:acc_level_2}\\
  \dot{p}_t &= \tau z_t\label{eq:vel_level_2} \\
  p_t &= \dot{p}_t \Delta t + p_{t-1}
\end{align}
where $\Delta t$ is the integration step size, and $p_t$, $\dot{p}_t$
and $\ddot{p}_t$ are position velocity and acceleration at time step
$t$.

%
by plugging in Equations~\ref{eq:acc_level_1},~\ref{eq:vel_level_1}
into Equations~\ref{eq:acc_level_2},~\ref{eq:vel_level_2} and setting
$\dot{z}_t = \frac{1}{\tau} \ddot{p}_t$ and
$z_t = \frac{1}{\tau} \dot{p}_t$ this can be reduced to
\begin{align}
  \ddot{p}_t &= \tau^2 (\alpha_z(\beta_z(g-p_{t-1}) - \frac{1}{\tau} \dot{p}_{t-1}) +f )\\
  \dot{p}_t &= \ddot{p}_t \Delta t + \dot{p}_{t-1} \\
  p_t &= \dot{p}_t \Delta t + p_{t-1}
\end{align}
By collecting $p_t$, $\dot{p}_t$ and $\ddot{p}_t$ into state
$\vs_t= \begin{pmatrix} \ddot{p}_t & \dot{p}_t & p_t \end{pmatrix}^T$
we we can summarize this linear system of equations as
\begin{align}
\vs_t = \mA \vs_{t-1} + \mB u_t
\end{align}
where the control input is given as $u_t = \alpha_z \beta_z g + s f_t$
and the transition matrix $\mA$ and control matrix $\mB$ are given as
$$\mA = \begin{pmatrix}0 & -\alpha_z\tau & -\alpha_z\beta_z \tau^2  \\ \Delta t & 1.0 & 0\\0 & \Delta t & 1.0 \end{pmatrix} \; \; \text{and} \; \;
\mB = \begin{pmatrix} 1 \\ 0 \\ 0 \end{pmatrix}$$

We like to account for two sources of uncertainty: transition noise,
modeling any uncertainty of transitioning from one state to the next
one; and observation noise, modeling noisy sensory measurements.
Next we show how to incorporate both of these to arrive at the full
probabilistic model.
A standard approach to modeling transition noise would be to assume
additive, zero mean, Gaussian noise, eg
\begin{equation}
\vs_t =  \mA\vs_{t-1} + \mB u_{t-1} + \mathbf{\epsilon}
\end{equation}
with $\mathbf{\epsilon} = \N(\mathbf{\epsilon} \g 0, \mQ)$ which would
create a time-independent transition uncertainty. Here, however, we
would like to model the transition uncertainty as a function of how
certain we are about our non-linear forcing term $f$. Remember, $f$
represents the shape of the motion primitive and is typically trained
via imitation learning. Assuming that we several demonstrations to
learn from, we can estimate the mean forcing term $f$ from the
demonstrations, but also the hvariance, using Bayesian regression.
Thus, for now, we assume that we have a predictive distribution over
$f_t$, the non-linear forcing term at time step $t$
\begin{align}
f_t \sim \N(f_t \g \mu_{f_t}, \sigma^2_t)
\end{align}
with mean $\mu_{f_t}$ and variance $\sigma^2_{f_t}t$. The details of
deriving this distribution are given in the next
Subsection~\ref{sec:learning_prob_dmps}.
Assuming $f_t$ to be drawn from a Gaussian distribution, automatically
implies that the hidden state $\vs_t$ is also Gaussian distributed
\begin{align}
  \vs_t \sim \N(\vs_t \g \mA \vs_{t-1} + \mB (\alpha_z \beta_z g + s \mu_{f_t}), Q_t)
\end{align}
where $Q_t = s^2 \mB \sigma^2_{f_t} \mB^T$. Thus, if the variance of
the distribution over $f_t$ is time-dependent, so is the state
transition noise.
Finally, we also want to be able to include noisy sensor measurements
in our model. Thus, we assume that we receive observations $o_t$ that
are a function of the hidden state, corrupted by zero mean Gaussian
noise
\begin{align}
o_t \sim h(\vs_t) + \N(v \g 0, R)
\end{align}
A simple example is for instance, the feedback on the actual position
of the system. In this case, the observation function would be
$h(\vs_t)= \mC \vs_t$. with observation matrix
$\mC = \begin{pmatrix} 0 & 0 & 1 \end{pmatrix}^T$.
Finally, putting it all together, our probabilistic formulation takes
to form of a controlled linear dynamical system, with time dependent
transition noise, and time-independent observation noise:
\begin{eqnarray*}
\vs_t &=&  \mA\vs_{t-1} + \mB (\alpha_z \beta_z g + s \mu_{f_t}) + \mathbf{\epsilon_t}\\
o_t &=& h( \vs_{t} ) + v
\end{eqnarray*}
with
$\mathbf{\epsilon}_t \sim \N(\epsilon_t \g 0, s^2 \mB \sigma^2_{f_t}
\mB^T)$ and $v \sim \N(v \g 0, R)$.
Note, for clarity, we have derived the probabilistic model for a
specific DMP variant \cite{Ijspeert2013}. However, the same procedure
can be followed to arrive at a dynamic graphical model representation
for other variants of motion primitives, such as
\cite{pastor2009learning}.
\subsection{Learning Probabilistic DMPs}\label{sec:learning_prob_dmps}
Probabilistic Dynamic Movement Primitives can be learned through
imitation learning similar to regular DMPs. Given $K$ demonstrations
$\mat{P}^k_\text{demo}$ of the same motion primitive, it is possible
to estimate the distribution over the non-linear function term $f$ via
Bayesian regression \cite{Bishop2006}. 

The standard approach to estimating the noise covariance $\mR$ of a
linear dynamical system is based on the
\emph{expectation-maximization}(EM) procedure. The EM algorithm
iterates between estimating the posterior of the hidden state of the
motion primitive, and maximizing the expected complete log likelihood
with respect to the parameter of interest. The complete data log
likelihood for $K$ demonstrations is given by:
\begin{align}
  \ln p(\mat{P},\mS |\tau,g)&= \sum_{k=1}^K \sum_{t=1}^T \ln \N(\vec{o}^k_t \g \mC \vs_t^k,\mR)+ \ln \N(\vs_1^k \g 0, \mQ_0) \notag\\ 
  & \; \;  + \sum_{k=1}^K \sum_{t=2}^T \ln \N(\vs_t^k | \mA \vs_{t-1}^k + \mB u^k_{t-1},\mQ_t)
\end{align}
Taking the expectation of the complete-data log likelihood with
respect to the posterior $p(\mS | \mat{P}, \theta^{old})$ defines the
function 
\begin{eqnarray*}
Q(\theta,\theta^{old}) = \mathbb{E}_{\mS| \theta^{old}}\big[ \ln p(\mat{P},\mS | \theta ) \big ],
\end{eqnarray*}
which we maximize with respect to parameters $\theta$. The updates for
$\mR$ can now be derived by setting the derivative of this function to
$0$, and then solve for $\mR$ which can be done analytically.
This completes the learning of a probabilistic dynamic movement
primitive, where the parameters required to fully describe the
probabilistic representation are
$\theta_\text{primitive} = \{\mu_w, \Sigma_w, \alpha, \beta, \mR \}$.

\subsection{Executing a probabilistic DMP}
\begin{figure}[t]
  \begin{footnotesize}
  \center
\begin{algorithmic}[1]
  \Procedure{Rollout}{$\vw$, $\mQ$, $\mR$}
  \For{$t=1:T$}
  \State $\mu_\text{p, t} = A \mu_\text{u, t-1} + B u_{t-1}$
  \State $\mat{V}_\text{p, t} = \mA \mat{V}_\text{u, t-1} \mA^T + \mQ$
  \EndFor
  \EndProcedure
\end{algorithmic}
\vspace{-0.3cm}
\caption{\small Probabilistic DMP: rollout}\label{alg:rollout}
  \end{footnotesize}
\end{figure}
\begin{figure}[t]
  \center
  \begin{small}
\begin{algorithmic}[1]
  \Procedure{ExecuteAndMonitor}{$\vw$, $\mQ$, $\mR$}
  \For{$t=1:T$}
  \State $\mu_\text{p, t} = A \mu_\text{u, t-1} + B u_{t-1}$
  \State $\mat{V}_\text{p, t} = \mA \mat{V}_\text{u, t-1} \mA^T + \mQ$
  \State $\mat{S} = \mC \mat{V}_\text{p, t} \mC^T + \mR$
  \State $K = \mat{V}_\text{p, t} \mC^T \mat{S}^{-1}$
  \State $\mu_\text{u, t} = \mu_\text{p, t} + K (o_t - \mC^T \mu_\text{p, t})$
  \State $\mat{V}_\text{u, t} = \mat{V}_\text{p, t} - K \mC \mat{V}_\text{p, t}$
  \EndFor
  \State \textbf{return} loglik 
  \EndProcedure
\end{algorithmic}
\vspace{-0.3cm}
\caption{\small Probabilistic DMP: executing and tracking a reference signal}\label{alg:forward_pass}
  \end{small}
\vspace{-0.3cm}
\end{figure}
Dynamic Movement Primitives are typically used to generate desired
trajectories that a controller is expected to track. Note, the hidden
state of the probabilistic DMP is given by
$\vs_t = \begin{pmatrix} \ddot{p}_t & \dot{p}_t & p_t \end{pmatrix}$
Pure feedforward trajectory generation is achieved by initializing the
linear dynamical system with the task parameters (goal, start and
duration of the motion) and then simply unrolling the probabilistic
model (see Algorithm~\ref{alg:rollout}) - this will generate exactly
the same desired trajectory as a standard DMP, including uncertainty
estimates.
As discussed above, we can also consider noisy sensor feedback when
executing a probabilistic DMP. Instead of simply forward predicting
the hidden state, inference is performed to estimate the hidden state
distribution. For linear dynamical systems this inference process is
widely known as Kalman filtering. In order to do so, we need to
formulate how the hidden state $\vs_t$ generates the chosen sensor
measurements $o_t$. This is done by defining the observation
function $h(\vs_t)$ that creates the reference signal we want to track
by transforming the hidden state.
\begin{figure*}[ht]
  \centering
  \includegraphics[width=0.295\linewidth]{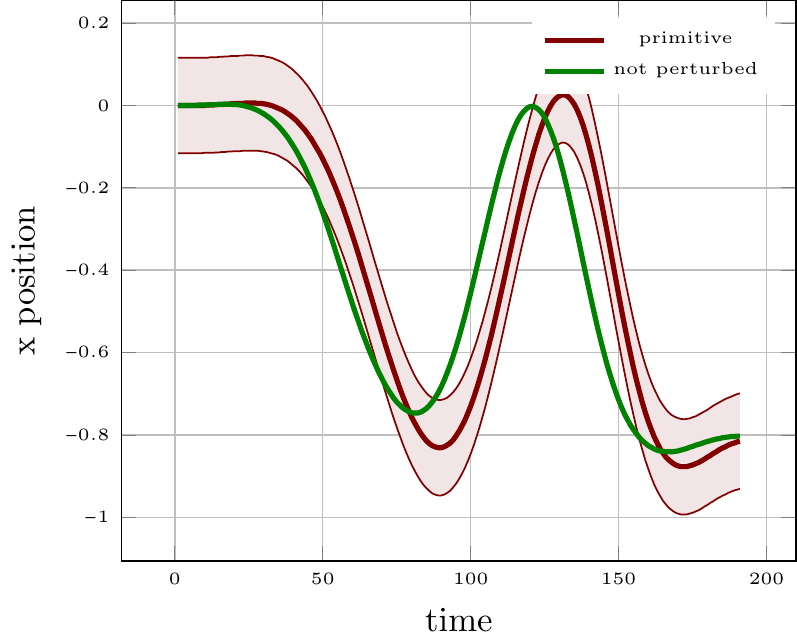}%
  \includegraphics[width=0.295\linewidth]{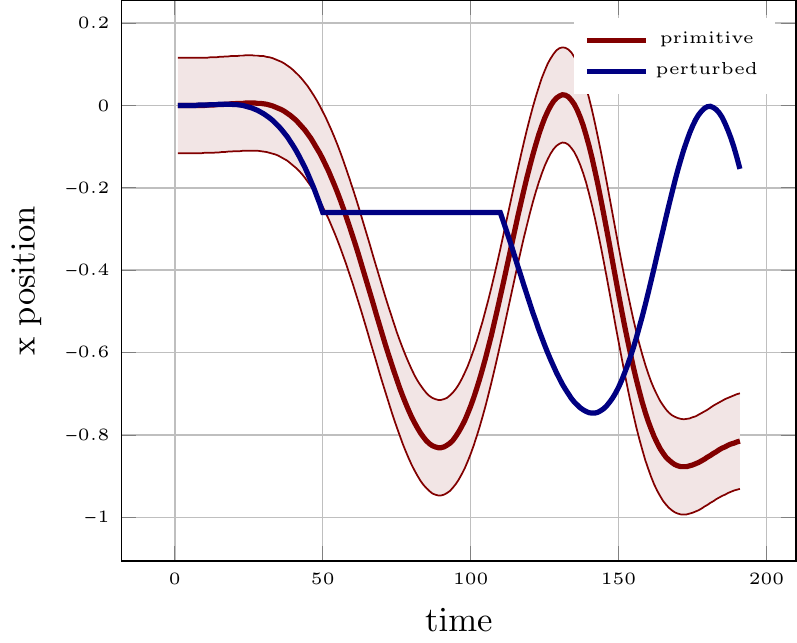}%
  \includegraphics[width=0.295\linewidth]{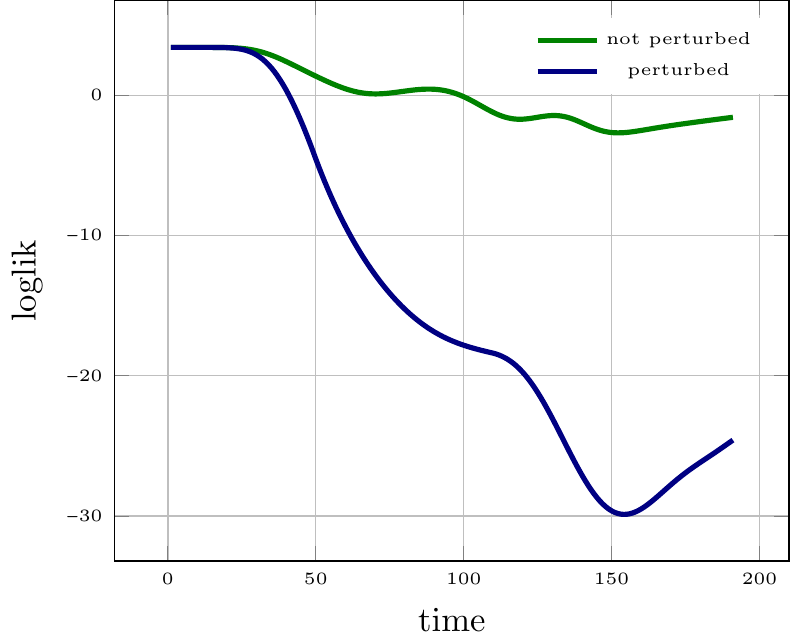}%
  \caption{\small Failure detection illustration: In the (left) and (middle)
    plot the mean trajectory of a learned primitive with the variance
    at each time step. Additionally: (left) the green trajectory
    illustrates a non-perturbed observed trajectory that slightly
    differs from the mean. (middle) The same trajectory but perturbed.
    (right) The log-likelihood of both observed trajectories
    calculated online while trajectories were
    unrolled.}\label{fig:failure_detection}%
    \vspace{-0.4cm}
\end{figure*}
Assuming, we observe the actual position of the system, this leads to
following inference steps: At time step $t$, the system feedforward
predicts the new desired (hidden) state $\mu_{p, t}$
\begin{align}
  \mu_{p, t} = \mA \mu_{t-1} + \mB u_{t-1}
\end{align}
then the Kalman innovation update, adds a term proportional to how
much the reference signal deviates from the observed sensor feedback
\begin{align}
\mu_{t} = \mu_{p, t} + K (o_{t} - h(\vs_t)).
\end{align}
Thus at time step $t+1$ the desired behavior of the DMP is online modulated
\begin{align}
  \mu_{p, t+1} &= \mA (\mu_{p, t} + K (o_{t} - h( \vs_t)) + \mB u_{t}\\
            &= \mA \mu_{p, t} + \mB u_{t} + \mA K (o_{t} - h( \vs_t))
\end{align}
Note, the similarity of this update to the online modulation performed
in \cite{Pastor2011}. To summarize, the execution of a probabilistic
DMP with noisy measurement observations is performed via Kalman
filtering (see Algorithm~\ref{alg:forward_pass}) and automatically
leads to online adaptation of desired behaviors to account for
disturbances.
\subsection{Summary}
In summary, the main characteristics of our probabilistic DMP
representation are:
\begin{itemize}
\item It is a probabilistic model that keep accelerations
  part of the hidden state - this allows to execute probabilistic DMPs
  just as regular DMPs - where a rollout of the hidden states produces
  desired accelerations, velocities and positions.
\item The non-linear forcing term is modeled probabilistically with
  phase-dependent variability - creating a phase-dependent transition
  covariance in the linear dynamical system view.
\item A reference signal can be tracked as part of the probabilistic
  model - which creates a principled way of modulating the desired
  behavior online, when the sensor feedback deviates from the reference signal.
\end{itemize}
\section{Failure Detection with Probabilistic DMPs}
\label{sec:applications}
Besides the insight presented above, other benefits of this
probabilistic formulation exist. For instance, assuming we have
learned a probabilistic representation for a motion primitive
$\{ \mu_w, \Sigma_w, \alpha, \beta, \mR \}$ we can perform online
failure detection: While executing the motion
primitive using Algorithm~\ref{alg:forward_pass}, we can utilize the probabilistic
formulation to continuously monitor how likely it is that this motion
primitive is generating the measured actual state of the system. %

We illustrate this application in Figure~\ref{fig:failure_detection}.
This illustration shows a 1D primitive being executed, first
perturbation free (right), and then we artificially hold the movement
such that the primitive is not continued to executed but the
probabilistic model expects it to (middle). On the left hand side we
see how the likelihood values evolve during movement execution. Note,
how once we artificially hold the movement the likelihood value
significantly drops. Thus, we can use this likelihood measure to
detect execution failures.

For initial quantitative evaluation purposes we recorded a dataset of
2D trajectories of letters with a digitizing tablet. All letters that
are easily written with one stroke have been recorded, a total of 22.
Each of this letter is meant to represent a movement primitive. To
learn a probabilistic representation per primitive we collected $10$
training demonstrations and an additional $10$ demonstrations for
testing purposes.

Once a probabilistic representation of a motion primitive has been
learned, we measure the minimum likelihood of each training
demonstration given the learned model parameters, and store that value
with the parameters.
Throughout the execution of a motion primitive the loglikelihood value
might increase or decrease depending on how much variation from the
mean we observe. The challenge is thus to detect natural variation
from a perturbation and/or failure. Here we simply classify an
execution as failed, if at any point during execution the
loglikelihood values drops below 2 times the minimum reported value
for that primitive. On our data this works very well, such that of all
$22*10=220$ test cases only 2 test cases where classified as failed when not
perturbed. When we perturb the simulated execution of each test case
by artificially blocking the execution (as illustrated in
Figure~\ref{fig:failure_detection}), then all $220$ test cases are
classified as failed.
\section{Conclusions}
We have presented a probabilistic model for dynamic movement
primitives. Inference in this graphical model is equivalent to Kalman
filtering, and when performing inference a feedback term is
automatically added to the DMP trajectory generation process. Besides
this insight, we have shown the potential of probabilistic DMPs on the
application of failure detection. Future work will explore and
evaluate the potential impact of these probabilistic representation in
more detail.

\bibliographystyle{aaai}
\bibliography{references}
\end{document}